\documentclass[letterpaper]{article} % DO NOT CHANGE THIS
\usepackage{aaai24}  % DO NOT CHANGE THIS
\usepackage{times}  % DO NOT CHANGE THIS
\usepackage{helvet}  % DO NOT CHANGE THIS
\usepackage{courier}  % DO NOT CHANGE THIS
\usepackage[hyphens]{url}  % DO NOT CHANGE THIS
\usepackage{graphicx} % DO NOT CHANGE THIS
\usepackage{natbib}  % DO NOT CHANGE THIS AND DO NOT ADD ANY OPTIONS TO IT
\usepackage{caption} % DO NOT CHANGE THIS AND DO NOT ADD ANY OPTIONS TO IT
\usepackage{newfloat}
\usepackage{listings}
\usepackage{booktabs}
\usepackage{multirow}
\usepackage{graphicx}
\usepackage[ruled,vlined]{algorithm2e}
\usepackage{amsmath}
\usepackage{hyperref}
\usepackage{bbding}
\usepackage{subcaption}
\usepackage{amssymb}% http://ctan.org/pkg/amssymb
\usepackage{pifont}% http://ctan.org/pkg/pifont
\newcommand{\cmark}{\ding{51}}%
\newcommand{\xmark}{\ding{55}}%

\newcommand{\bw}{\mathbf{w}}
\newcommand{\bs}{\mathbf{s}}

\newcommand{\CC}{\mathcal{C}}
\newcommand{\NN}{\mathcal{N}}
\newcommand{\DD}{\mathcal{D}}
\newcommand{\LL}{\mathcal{L}}
\DeclareCaptionStyle{ruled}{labelfont=normalfont,labelsep=colon,strut=off} % DO NOT CHANGE THIS
\title{Federated Learning with Extremely Noisy Clients via Negative Distillation}
\author{
    Yang Lu \textsuperscript{\rm 1,\rm 2},
    Lin Chen \textsuperscript{\rm 1,\rm 2},
    Yonggang Zhang \textsuperscript{\rm 3},
    Yiliang Zhang \textsuperscript{\rm 1,\rm 2},\\
    Bo Han \textsuperscript{\rm 3},
    Yiu-ming Cheung \textsuperscript{\rm 3},
    Hanzi Wang \textsuperscript{\rm 1,\rm 2}\thanks{Corresponding author.}
}

\affiliations{
   \textsuperscript{\rm 1}Fujian Key Laboratory of Sensing and Computing for Smart City, School of Informatics, Xiamen University, China\\
    \textsuperscript{\rm 2}Key Laboratory of Multimedia Trusted Perception and Efficient Computing,\\ Ministry of Education of China, Xiamen University, China\\
    \textsuperscript{\rm 3}Department of Computer Science, Hong Kong Baptist University, Hong Kong, China\\
    luyang@xmu.edu.cn,
    chenlin191209@163.com,
    ylzhangcs@hotmail.com,
    hanzi.wang@xmu.edu.cn,\\
    csygzhang@comp.hkbu.edu.hk,
    bhanml@comp.hkbu.edu.hk,
    ymc@comp.hkbu.edu.hk
}

\usepackage{bibentry}
\begin{document}

\maketitle

\begin{abstract}
Federated learning (FL) has shown remarkable success in cooperatively training deep models, while typically struggling with noisy labels. Advanced works propose to tackle label noise by a re-weighting strategy with a strong assumption, i.e., mild label noise. However, it may be violated in many real-world FL scenarios because of highly contaminated clients, resulting in extreme noise ratios, e.g., $>$90\%. To tackle extremely noisy clients, we study the robustness of the re-weighting strategy, showing a pessimistic conclusion: minimizing the weight of clients trained over noisy data outperforms re-weighting strategies. To leverage models trained on noisy clients, we propose a novel approach, called \emph{ne}gative \emph{d}istillation (FedNed). FedNed first identifies noisy clients and employs rather than discards the noisy clients in a knowledge distillation manner. In particular, clients identified as noisy ones are required to train models using noisy labels and pseudo-labels obtained by global models. The model trained on noisy labels serves as a `bad teacher' in knowledge distillation, aiming to decrease the risk of providing incorrect information. Meanwhile, the model trained on pseudo-labels is involved in model aggregation if not identified as a noisy client. Consequently, through pseudo-labeling, FedNed gradually increases the trustworthiness of models trained on noisy clients, while leveraging all clients for model aggregation through negative distillation. To verify the efficacy of FedNed, we conduct extensive experiments under various settings, demonstrating that FedNed can consistently outperform baselines and achieve state-of-the-art performance.
Our code is available at \href{https://github.com/linChen99/FedNed}{Github}.
\end{abstract}

\section{Introduction}
The rise of federated learning (FL) benefits from its capacity for large-scale distributed model training in a data-preserving manner \cite{kairouz2021advances}. 
The server aggregates client models to produce a global model and sends it back for subsequent training. 
When the sample annotation is accurate, the global model can generally exhibit promising performance, even when the data is somehow non-IID distributed \cite{ma2022state}.
Another challenge in FL is the label-noise problem. 
Usually, as each client collects and annotates the data by itself, the inaccurate annotation in each client may be with different degrees \cite{xu2022fedcorr}. 
Different from the label-noise learning in a batch setting, the server in FL needs to judge the degree of label noise for each client before model aggregation, because the server has no information about which client has label noise.

\begin{figure}
    \centering
    \includegraphics[width=.9\linewidth]{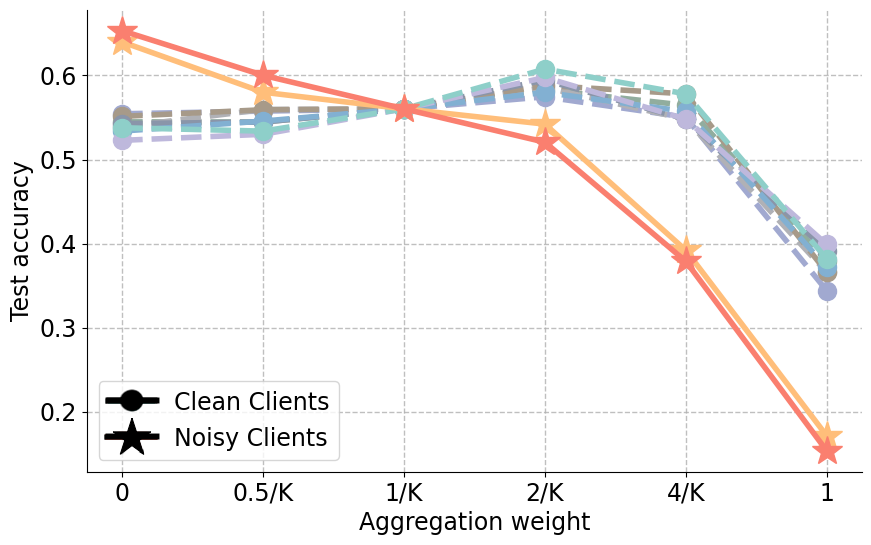}
    \caption{The test accuracy of the global model by controlling the weight of a single client model. We set ten client models including eight clean ones (with a noise ratio of 0\%) and two extremely noisy ones (with a noise ratio of 99\%). $K$ is the total number of clients, which is ten in this example.}
    \label{motivation}
\end{figure}

Many efforts have been devoted to alleviating the label-noise issue in FL~\cite{liang2023fednoisy,yang2021client,chen2020focus,wang2022fednoil,yang2022robust,kim2022fedrn,xu2022fedcorr,wu2023fednoro}.
Advanced works have shown the promising benefits of assigning different weights to each client during the model aggregation process. The intuition behind these methods is that all client models have the potential to contribute to the global model by model aggregation, highlighting the importance of aggregation weights assigned for each client. The basic intuition is built upon a strong assumption that the label noise on each client is relatively mild. 
Namely, models trained on mild noise can benefit global models by aggregation.

However, the strong assumption could be violated in many practical FL scenarios. For instance, some clients may be highly contaminated with noise ratios exceeding $90\%$, due to unintentional mislabeling or deliberate data poisoning. These clients are referred to as `extremely noisy' clients in the context of this work. Consequently, models trained on extremely noisy clients may perform differently on the same dataset, causing global models to degrade via model aggregation. We assign different weights used for aggregation to illustrate the negative impacts of models trained on extremely noisy clients, as depicted in Figure~\ref{motivation}. These experiments show that the performance of the global model varies with the weight assigned to a client model. Specifically, for each line plot, we merely change the weight for one client model, while keeping weights for the left clients equally \footnote{The sum of all weights is 1.}. Our results show that a) models trained on clean clients contribute to the global model, and b) models trained on extremely noisy clients lead to severe performance degradation. Namely, we should discard models trained on noisy clients, i.e., assigning $0$ weights to these models rather than weighing them with an arbitrarily small weight. Therefore, discarding noisy clients is preferred over re-weighting clients for the model aggregation process. However, the discarding strategy goes against the intention of FL.

In this work, we propose Federated learning via Negative distillation (FedNed) to deal with the extreme-noise problem. FedNed first identifies the client models with extreme label noise by model prediction uncertainty~\cite{gal2016dropout}, since uncertainty is widely used to measure whether a model can be trusted~\cite{jiang2018trust}. Then, rather than directly discarding them, FedNed utilizes them through a novel strategy called negative distillation. In negative distillation, these client models trained on extremely noisy data act as `bad teachers' when updating the global model. FedNed keeps the global model's prediction different from that of the extremely noisy client models, which shares the same spirit with negative learning~\cite{kim2019nlnl}, i.e., reducing the risk of providing incorrect information. As a result, negative distillation produces an even better global model than the one aggregated by only using client models. Extensive experiments verify the effectiveness of the proposed method on the environment of clients with extremely noisy-labeled data.

Our main contributions are summarized as follows:
\begin{itemize}
\item We reveal the severe impacts induced by extremely noisy clients, posing challenges to existing methods. Specifically, involving models trained on extremely noisy clients causes performance degradation of global models.

\item We propose a novel method called FedNed to tackle FL with extremely noisy clients. In FedNed, the key idea called negative distillation is proposed to encourage the global model's prediction to be dissimilar to that of noisy models. A new local optimization strategy is subsequently adopted for identified extremely noisy clients.

\item We conduct comprehensive experiments to verify the efficacy of FedNed on benchmarks with extremely noisy clients, demonstrating that
FedNed significantly and consistently outperforms state-of-the-art methods. 
\end{itemize}

\section{Related Work}
We first summarize the advancements achieved in the domains of federated learning (FL) and label-noise learning.
Then, we summarize the recent work of the joint problem: label-noise learning in the FL environment.

\subsection{Federated Learning with Data Heterogeneity}
Since the seminal work FedAvg~\cite{mcmahan2017communication} was proposed, the landscape of FL research has predominantly revolved around addressing challenges of data heterogeneity~\cite{ma2022state}, where data distributions shift with clients, i.e., non-IID data. 

Advanced works have achieved outstanding improvements through various approaches, under an assumption that client data are noise free. FedProx~\cite{li2020federated} introduces a regularization mechanism into the local training process by employing a proximal term, effectively enhancing convergence behavior. 
SCAFFOLD~\cite{karimireddy2020SCAFFOLD} mitigates client drift by incorporating supplementary control variates, ensuring stable convergence.
FedDyn~\cite{durmus2021federated} dynamically regulates the training process of neural network models across devices, allowing for efficient training while remaining robust to diverse scenarios.
Recently, FedNH~\cite{dai2023tackling} marks an important stride by enhancing the efficacy of local models in both personalization and generalization.
It is achieved through the incorporation of uniformity and class semantics in class prototypes, thereby improving overall model stability and effectiveness across diverse clients.
FedNP \cite{wu2023FedNP} efficiently estimates the inaccessible ground-truth global data distribution using a probabilistic neural network, mitigating performance degradation induced by data heterogeneity. Advanced works propose to share privacy-free data among clients to tackle data heterogeneity~\cite{tang2022virtual,yang2023fedfed}, achieving promising performance.

\subsection{Label-Noise Learning}
In numerous real-world scenarios, data annotation often gives rise to the challenge of noisy labels. A considerable number of methods for label-noise learning can be categorized into the following groups~\cite{song2022learning}: sample selection~\cite{yao2021jo, karim2022unicon}, loss function adjustment~\cite{ghosh2017robust, shu2019meta}, regularization~\cite{xia2020robust, lukasik2020does}, and robust model architecture~\cite{han2018co, han2018masking}. Among these works, the line of sample selection is the most related approach, as we perform model selection to defy label noise.

Early methods primarily rely on the trick of small risk (or loss), sharing the same spirit with model selection using uncertainty~\cite{jiang2018trust}. For instance, co-teaching~\cite{han2018co} adopts sample selection by two distinct models, wherein the clean samples selected by one model are used to train another. Similarly, DivideMix~\cite{li2019dividemix} effectively employs the small loss trick to select clean samples, subsequently integrating semi-supervised learning by treating the unselected samples as unlabeled. Recently, contrastive learning approaches have been involved in sample selection.
Jo-SRC~\cite{yao2021jo} utilizes contrastive learning to estimate the likelihood of sample cleanliness or out-of-distribution by training the network with dual predictions and introducing a joint loss with consistency regularization to improve model generalization. Unicon~\cite{karim2022unicon} employs a uniform selection mechanism, coupled with contrastive learning, to tackle imbalanced sample selection and prevent the memorization of noisy labels.

\begin{figure*}[!t]
    \centering
    \includegraphics[width=.9\linewidth]{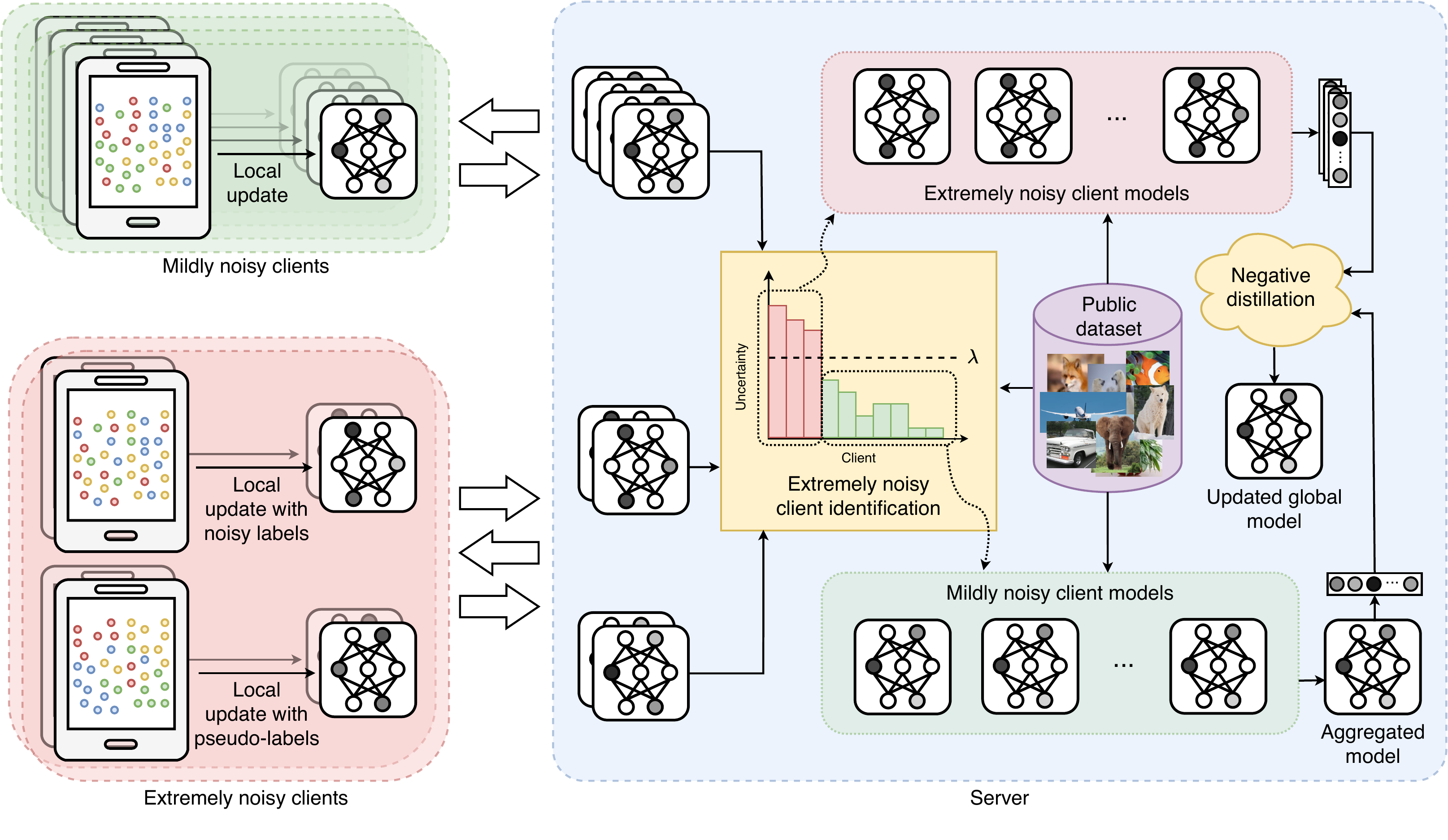}
    \caption{The architecture overview of the proposed FedNed. In each round, the server identifies the mildly noisy (MN) and extremely noisy (EN) client models via MC dropout and prediction uncertainty. Negative distillation is then utilized to incorporate EN client models for a better global model.}
    \label{model_figure}
\end{figure*}

\subsection{Federated Learning on Noisy-Labeled Data}\label{section_FL_noise}
The management of noisy-labeled local data from diverse clients poses a novel challenge within the field of FL \cite{liang2023fednoisy}. Specifically, advanced FL methods for data heterogeneity typically struggle with noisy labels, while traditional label-noise learning approaches are no longer robust when facing distributed data.

One straightforward approach is to reserve clean data on the server to identify noisy clients.
Among these methods, quantifying each client's noise ratio is used to identify low-noise ratio clients~\cite{yang2021client}. The server can subsequently aggregate the client models based on the ranking of estimated noise ratios. Similarly, FOCUS~\cite{chen2020focus} assigns different weights to clients based on the credibility of their local data.
However, these methods make a strong assumption that the server holds clean labeled data. 

Advanced works make a great attempt to weaken the assumption when estimating noise rate. For instance, FedNoiL~\cite{wang2022fednoil} utilizes prediction confidence to estimate the noise ratio, which is then normalized to weight client models during model aggregation. FedCorr~\cite{xu2022fedcorr} adopts local intrinsic dimension to differentiate between clean and noisy clients, utilizing the estimated noise rate as a regularization coefficient to constrain model updates. FedNoRo~\cite{wu2023fednoro} initially identifies noisy clients through normalized local losses, followed by distance-aware model aggregation in the second stage. RoFL~\cite{yang2022robust} assesses distances between prototypes to facilitate model aggregation. FedRN~\cite{kim2022fedrn} identifies some reliable neighbors and employs their mixture model to enhance clean sample selection. However, the intuition behind these methods is that all client models can contribute to the global model, which may no longer hold under extremely noisy scenarios.

\section{Proposed Method}
In this section, we detail how the proposed method FedNed endows FL with robustness against extremely noisy clients.

\subsection{Problem Definition}
In the typical FL environment~\cite{mcmahan2017communication}, a collection of $K$ clients collaborates with a central server to train a global model. Each client collects its local dataset, denoted as $\mathcal{D}_k$, which may contain noisy-labeled samples with an unknown noise ratio. Regarding the noise ratio, existing works assume that it would never surpass a certain threshold, i.e., mild label noise. Many practical scenarios highlight the importance and urgency of relaxing the strong assumption to that of extreme label noise.

In our work, the noise ratio ranges from $0\%$ to $100\%$, where clients are categorized as `extremely noisy clients' if their uncertainties exceed a certain threshold. Our focus is mainly on these extremely noisy clients, since this complex scenario is rarely discussed in the literature.
\subsection{Algorithm Overview}
The overall training process of FedNed is illustrated in Figure~\ref{model_figure}. 
The server identifies the extremely noisy clients in each round (c.f. Section \ref{step_identification}) and excludes their uploaded models during model aggregation.
Subsequently, a novel negative distillation procedure (c.f. Section \ref{step_ND}) is employed to further enhance the global model's performance by incorporating information from the extremely noisy models. On the server, we utilize a public dataset $\DD_U$ for both the identification step and the negative distillation step. To address the privacy issue in FL, $\DD_U$ does not need to be part of the training data and can be a public dataset without label annotation. Local training on each client also takes into account the identification result on the server in the previous round.
If a client is identified as extremely noisy, an additional local model is updated and uploaded (c.f. Section \ref{step_local}).

\subsection{Identification of Extremely Noisy Clients}\label{step_identification}
The first step is to identify the client models trained with extreme label noise. In each communication round, the server randomly selects a subset of clients denoted as $\mathcal{A}^t$. Subsequently, the selected clients are divided into two categories: mildly noisy (MN) clients indexed by $\CC^{t}$ and extremely noisy (EN) clients indexed by $\NN^{t}$. EN clients are updated on local extremely noisy-labeled data.

Following previous work~\cite{jiang2018trust}, we employ model uncertainty~\cite{gal2016dropout} to identify models with high risk, i.e., trained with noisy-labeled data. Specifically, client models with high uncertainty are regarded as EN clients. Denoting $\bw_k^{t}$ as the parameter of local model at client $k$ for the $t$-th round, the probability of model for class $c$ can be calculated by:
\begin{align}\label{class_uncertainty}
p(y=c|\mathbf{x}, \DD_k)&:=\int p(y=c|\mathbf{x}, \bw)p(\bw|\DD_k)d\bw\\
&\approx \frac{1}{T}\sum_{t=1}^Tp(y=c|\mathbf{x}, \widehat\bw_t),
\end{align}
where $\mathbf{x} \in \DD_U$ is input data with its label $y$, $p(y=c|\mathbf{x}, \bw)$ stands for the probability label $c$ predicted by model with parameter $\bw$, $p(\bw|\DD_k)$ represents the distribution of applying Dropout operation to models trained on local data $\DD_k$ during inference, $T$ is the times to perform inference for each sample, and $\widehat\bw_t$ denotes parameters sampled from $p(\bw|\DD_k)$. Built upon $p(y=c|\mathbf{x}, \DD_k)$, we calculate the uncertainty $U_k$ for client $k$ by averaging over all samples and classes:
\begin{align}\label{entropy}
    \frac{-1}{C\cdot|\DD_U|}\sum_{c=1}^C\sum_{\mathbf{x}\in D_U}
    p(y=c|\mathbf{x}, \DD_k)\log p(y=c|\mathbf{x}, \DD_k).
\end{align}
We select the clients whose uncertainty $U_k$ is greater than a threshold $\lambda$ as the EN clients.

It is worthwhile to note that client models identified as noisy are used to perform model aggregation in previous works~\cite{wu2023fednoro,wang2022fednoil}, while we identify noisy clients to perform negative distillation.

\subsection{Negative Distillation}\label{step_ND}
Building upon the observation and analysis presented in Figure \ref{motivation}, the integration of EN clients into model aggregation with arbitrary weights emerges as detrimental to the generalization performance of the global model.
Opting for a straightforward solution that involves discarding the EN clients offers a protective measure to maintain the integrity of the global model. 
However, these clients could potentially hold information beneficial to enhance the global model.
Apparently, the EN clients are trained on the local datasets with extremely noisy labels, leading to their diminished capacity for accurate predictions. 
Thus, leveraging the incorrect predictions generated by EN clients may compel the global model to diverge from their predictions. Avoiding incorrect predictions is widely used in negative learning, which aims to reduce the risk of providing incorrect information. In this paper, we implement this idea by the concept of negative distillation, making the global model's prediction diverge from those offered by the identified EN clients. Negative distillation is utilized to incorporate EN client models for a better global model.

Inspired by FedDF \cite{lin2020ensemble}, we leverage knowledge encoded in EN by knowledge. In contrast to FedDF's strategy of ensembling all client models for global model improvement, we consider the client models from EN clients as the `bad teacher'. The goal is to make the student model (e.g., the global model) remain distant from the `bad teacher'.
The server initializes the student model by aggregating solely the MN client models:
\begin{align}\label{model_aggregate}
\bs^{t+1}=\sum_{k\in \CC^{t}}\frac{N_k}{N^t}\bw_k^{t},
\end{align}
where $N_k$ is the number of training samples in client $k$ and $N^t=\sum_{k\in\CC^{t}}N_k$ is the total number of training samples of the selected MN clients in round $t$.

The initial student model is solely aggregated by MN client models, thereby preventing being affected by EN client models during model aggregation.
Subsequently, the student model is updated by optimizing the negative distillation loss function $\LL_{nd}$:
\begin{align}\label{negative_distillation}
    \frac{1}{|\NN^{t}| |\DD_U|}
    \sum_{k\in\NN^{t}, \mathbf{x}\in \DD_U}
    d\left[ f(\mathbf{x};\bs^{t+1}),
    g(\mathbf{x};\bw_k^{t+1})\right],
\end{align}
where $d(\mathbf{u}, \mathbf{v}):=KL\left[\sigma(\mathbf{u}), \sigma(\mathbf{v})\right]$ is the distance based on KL divergence with $\sigma(\cdot)$ the softmax activation function, $f(\mathbf{x};\bw)=\{f_1,f_2,...,f_C\}$ stands for the output probability vector, and $g(\mathbf{x};\bw):=\{f_1^{-1},f_2^{-1},...,f_C^{-1}\}$ are the reciprocals of the output.
$KL(\mathbf{u},\mathbf{v})=(\sigma(\mathbf{u}),\sigma(\mathbf{v}))$ is the distillation loss and $\sigma$ is the softmax function. 
This loss compels the student model's predictions to diverge from those of each EN client model.
The reciprocals of the output from EN client models could include knowledge from student models, given that these models tend to produce incorrect predictions consistently.
In this manner, the student model is enhanced by avoiding wrong knowledge from the `bad teacher'.
Finally, the updated student model is sent back to each client as the global model $\bw^{t+1}$. 

\begin{algorithm}[tb]
	\label{alg:algorithm}
    \caption{Federated Negative Distillation}
    \LinesNumbered
	\KwIn{$T_0$ is the number of warm-up rounds; $T$ is the total number of rounds; $\lambda$ is the threshold for selecting EN clients; and $K$ is the total number of clients.}
	\KwOut{The global model $\bw^{T}$ in round $T$ }
    Initialize local model $\bw_k^0$ for each client;\\
	\For{$t=1$ \KwTo $T$}{ 
		\tcp{Clients execute:}
        \For{$k=1,...,K$}{	
			Update local model $\bw_k^{t+1}$ by Eq. (\ref{client_normal_update});\\
			Send $\bw_k^{t+1}$ to the server;\\
            \If{$k\in \NN^t$ and $t>T_0$}{
                Update local model $\widehat\bw_k^{t+1}$ by Eq. (\ref{client_pseudo_update});\\
                Send $\widehat\bw_k^{t+1}$ to the server;
            }
		}
		\tcp{Server executes:}
        Randomly select a set of active clients $\mathcal{A}^t$;\\
        Select $\CC^{t}$ and $\NN^{t}$ by Eq.~(\ref{entropy});\\
		Aggregate local models to $\bs^{t+1}$ by Eq.~(\ref{model_aggregate});\\
        \If{$t>T_0$}{
            Update $\bw^{t+1}$ by Eq.~(\ref{negative_distillation});\\
        }
        \Else{
            $\bw^{t+1}\gets \bs^{t+1}$;\\
        }
		Send $\bw^{t+1}$ and $\NN^t$ to clients.\\
	}
\end{algorithm}

% \begin{algorithm}[!h]
% 	\caption{Federated Negative Distillation}
% 	\label{alg:algorithm}
%     \begin{algorithmic}[1]
% 	\Require $T_0$ is the number of warm-up rounds; $T$ is the total number of rounds; $\lambda$ is the threshold for selecting EN clients; and $K$ is the total number of clients.
% 	\Ensure The global model $\bw^{T}$ in round $T$ 
%     Initialize local model $\bw_k^0$ for each client;\\
    
% 	\FOR{$t=1$ \to $T$} 
% 		% \tcp{Clients execute:}
%         \FOR{$k=1,...,K$}	
% 			Update local model $\bw_k^{t+1}$ by Eq. (\ref{client_normal_update});\\
% 			Send $\bw_k^{t+1}$ to the server;\\
%             \IF{$k\in \NN^t$ and $t>T_0$}
%                 Update local model $\widehat\bw_k^{t+1}$ by Eq. (\ref{client_pseudo_update});\\
%                 Send $\widehat\bw_k^{t+1}$ to the server;
%             \ENDIF
%         \ENDFOR
%     \ENDFOR

%     % \tcp{Server executes:}
%     Randomly select a set of active clients $\mathcal{A}^t$;\\
%     Select $\CC^{t}$ and $\NN^{t}$ by Eq.~(\ref{entropy});\\
%     Aggregate local models to $\bs^{t+1}$ by Eq.~(\ref{model_aggregate});\\
%     \IF{$t>T_0$}
%         Update $\bw^{t+1}$ by Eq.~(\ref{negative_distillation});\\
%     \ElSE
%         $\bw^{t+1}\gets \bs^{t+1}$;\\
%     Send $\bw^{t+1}$ and $\NN^t$ to clients.\\
% 	\ENDIF
%  \end{algorithmic}
% \end{algorithm}

\begin{table*}[!t]
    \resizebox{\linewidth}{!}{
    \begin{tabular}{lcccccc|cccccc}            
		\toprule                               
		\textbf{Method}&                      
		\multicolumn{6}{c}{\textbf{CIFAR-10}}&          
		\multicolumn{6}{c}{\textbf{CIFAR-100}}\cr     
		\cmidrule{2-13}         
        Beta &\multicolumn{2}{c}{(0.1, 0.1)}
        &\multicolumn{2}{c}{(0.1, 0.3)}
        &\multicolumn{2}{c}{(0.3, 0.5)}
        &\multicolumn{2}{c}{(0.1, 0.1)}
        &\multicolumn{2}{c}{(0.1, 0.3)}
        &\multicolumn{2}{c}{(0.3, 0.5)}\\    
        \cmidrule{2-13} 
		Dirichlet &0.7 &10 &0.7 &10 &0.7 &10 &0.7 &10 &0.7 &10 &0.7 &10\\
		\midrule                                
        FedAvg&61.51&69.26&75.43&77.86&69.79&71.34&36.81&39.38&38.57&39.91&36.86&38.77\\
		FedProx&69.81&74.69&77.72&80.31&71.77&75.87&40.10&42.81&41.04&42.39&39.52&42.32\\
  		SCAFFOLD&64.07&68.42&75.96&76.83&70.21&73.41&39.64&41.19&40.14&40.72&40.11&41.45\\
		FedDyn&66.04&69.41&76.41&80.24&72.57&76.47&28.81&30.04&28.98&31.54&28.70&31.33\\
        RoFL&71.64&79.05&77.03&81.26&77.58&79.04&43.28&46.07&49.36&49.41&45.95&46.42\\
        FedCorr&74.10&78.35&81.91&\underline{85.10}&74.55&\underline{80.06}&40.24&44.33&\underline{52.76}&\underline{57.49}&\underline{47.03}&\textbf{51.23}\\
        FedNoRo&\underline{80.25}&\underline{80.63}&\underline{82.41}&84.11&\underline{77.67}&77.83&\underline{46.33}&\underline{47.15}&48.41&48.69&47.02&47.56\\
        
        FedNed (Ours)&\textbf{82.83}&\textbf{85.12}&\textbf{84.97}&\textbf{86.84}&\textbf{79.43}&\textbf{82.64}&\textbf{47.85}&\textbf{48.32}&\textbf{53.74}&\textbf{57.93}&\textbf{48.21}&\underline{49.68}\\
		\bottomrule                            
	\end{tabular}
 }
 \caption{Numerical comparison between the proposed FedNed and other FL methods with extremely noisy clients. The best results are highlighted in bold, while the second-best results are underlined.}
    \label{big_table}
\end{table*}

% \begin{table*}[!t]
% \centering
% 	\caption{Numerical comparison between the proposed FedNed and other FL methods with extremely noisy clients. The best results are highlighted in bold, while the second-best results are underlined.}
%     \begin{tabular}{lcccccc}            
% 		\toprule                               
% 		\textbf{Method}&                      
% 		\multicolumn{6}{c}{\textbf{CIFAR-10}}\\       
% 		\cmidrule{2-7}         
%         Beta &\multicolumn{2}{c}{(0.1, 0.1)}
%         &\multicolumn{2}{c}{(0.1, 0.3)}
%         &\multicolumn{2}{c}{(0.3, 0.5)}\\    
%         \cmidrule{2-7} 
% 		Dirichlet &0.7 &10 &0.7 &10 &0.7 &10 \\
% 		\midrule                                
%         FedAvg&61.51&69.26&75.43&77.86&69.79&71.34\\
% 		FedProx&69.81&74.69&77.72&80.31&71.77&75.87\\
%   		SCAFFOLD&64.07&68.42&75.96&76.83&70.21&73.41\\
% 		FedDyn&66.04&69.41&76.41&80.24&72.57&76.47\\
%         FedCorr&74.10&78.35&81.91&\underline{85.10}&74.55&\underline{80.06}\\
%         RoFL&71.64&79.05&77.03&81.26&77.58&79.04\\
%         FedNoRo&\underline{80.25}&\underline{80.63}&\underline{82.41}&84.11&\underline{77.67}&77.83\\
        
%         FedNed&\textbf{82.83}&\textbf{85.12}&\textbf{84.97}&\textbf{86.84}&\textbf{79.43}&\textbf{82.64}\\
% 		\bottomrule                            
% 	\end{tabular}
%     \label{big_table}
% \end{table*}

\subsection{Client-Side Training}\label{step_local}
Once the server has identified the EN clients in round $t$, the subsequent client training in round $t+1$ is tailored to address their unique characteristics.
Given that the data in EN clients is predominantly noisy, we employ a straightforward approach of training two distinct local models for each EN client.
The first local model is trained by discarding the noisy labels entirely and updated in an unsupervised manner.
This unsupervised local model is denoted as $\widehat\bw_k^{t+1}$, which is updated by:
\begin{align}\label{client_pseudo_update}
\widehat\bw_k^{t+1}\gets \bw_k^t-\eta\nabla_{\bw}\ell(\bw^t;\widehat\DD^k),
\end{align}
where $\widehat\DD^k$ is the local dataset with pseudo-labels assigned by the global model $\bw^{t}$ at round $t$.
The second local model continues to train on the original local dataset $\DD_k$:
\begin{align}\label{client_normal_update}
\bw_k^{t+1}\gets \bw_k^t-\eta\nabla_{\bw}\ell(\bw^t;\DD^k).
\end{align}
These two local models are sent to the server for global model updating, with the expectation that they contribute to the global model through both model aggregation and negative distillation, respectively.

In summary, during the local training round $t+1$, every client acquires an updated supervised model $\bw_k^{t+1}$ on its local dataset $\DD_k$, while only the EN clients acquire the additional unsupervised models $\widehat\bw_k^{t+1}$ trained on the pseudo-labeled local dataset $\widehat\DD_k$.
Hence, a total $k+|\NN_k^{t}|$ models are uploaded to the server in each round. 
Given the relatively small number of EN clients, the incurred communication cost remains manageable.
The ablation study in Section \ref{section_ablation} confirms the efficacy of this client-side training approach.

In addition, in order to prevent an MN client that being wrongly identified as an EN client during the early training phase, we conduct a warm-up training phase in the first few rounds.
During this phase, the server exclusively aggregates the MN clients without any intervention of the client training.
Algorithm 1 shows the entire training process for both clients and the server.
\section{Experiments}

\subsection{Experimental Settings}

\subsubsection{Datasets}

In the experiments, we adopt CIFAR-10 and CIFAR100~\cite{2009Learning} to verify the efficacy of the proposed method.
To accommodate the setting of FL with extremely noisy clients, we preprocess the training data through the following steps:
(1) First, we distribute the training data across each client in a non-IID manner.
Following the widely used strategy for generating non-IID clients \cite{yurochkin2019bayesian}, we utilize Dirichlet distribution with a parameter that controls the degree of data heterogeneity. 
(2) Then, we assign different noise ratios to individual clients, with a subset categorized as extremely noisy clients.
The noise ratio for each client is drawn from a Beta distribution $Beta(\alpha,\beta)$.
(3) We add label noise to each client based on the assigned noise ratio from the Beta distribution.
Due to data heterogeneity, we impose uniform noise on each client only for the classes represented within a client's local data.

For the public dataset $\DD_U$ on the server, we use different datasets from the clients' local data.
We use 128 images from CIFAR-100 as $\DD_U$ for training CIFAR-10, and 128 images from ImageNet~\cite{2015ImageNet} as $\DD_U$ for training CIFAR-100. All images are randomly selected from the dataset.
We simply use the official testing data split by the benchmark for global model testing.

\subsubsection{Training Details}
We use ResNet-18 for CIFAR-10, and ResNet-50 for CIFAR-100 as the base model. 
All the compared FL methods are implemented with the same model architecture. 
All experiments are run by PyTorch on two NVIDIA GeForce RTX 3090 GPUs. 
By default, we run 100 communication rounds to present the experimental results.
The total number of clients is set at 20, and an active client ratio $50\%$ is maintained in each round. 
For local training, the batch size is set at 32. 
We use SGD with a learning rate $0.05$ as the optimizer. The threshold $\lambda$ for the identification of EN client is set at $0.12$.

\subsection{Comparison with SOTA Methods}

We compare the proposed FedNed with two groups of methods. 
(1) FL baseline methods for data heterogeneity: FedAvg \cite{mcmahan2017communication}, FedProx \cite{li2020federated}, SCAFFOLD \cite{karimireddy2020SCAFFOLD} and FedDyn \cite{durmus2021federated}; and (2) Methods for FL with label noise: FedCorr \cite{xu2022fedcorr}, RoFL \cite{yang2022robust}, and FedNoRo \cite{wu2023fednoro}.
We also evaluate the robustness of the proposed method by varying the data distribution with different noise distributions (Beta) and data heterogeneity distributions (Dirichlet). The number of EN clients drawn from three Beta distributions (0.1, 0.1), (0.1, 0.3), and (0.3, 0.5) are about 5-6, 2-4, and 1-2, respectively, among a total number of 20 clients, where the Beta distribution (0.1, 0.3) has the overall minimum noise ratio.

Table \ref{big_table} shows the comparative results.
Evidently, label noise-oriented FL methods (e.g. FedCorr, RoFL, FedNoRo, and FedNed) consistently yield superior results compared to the FL baselines (e.g. FedAvg, FedProx, SCAFFOLD, and FedDyn), as the latter solely address the challenge of data heterogeneity. 
By comparing FedNed with methods for FL with label noise, FedNed notably outperforms the second-best method by approximately $2\%\hbox{-} 3\%$ on CIFAR-10, and generally exhibits better performance on CIFAR-100.
In addition, it can be observed that FedNed is less sensitive to the degree of data heterogeneity.
FedNed also demonstrates promising performance with different noise distributions.

\vspace{-2px}
\subsection{Model Validation}
We further delve into specific aspects related to how FedNed addresses the extreme noise challenge in FL.

\vspace{-2px}
\subsubsection{Ablation Study}\label{section_ablation}
We conduct an ablation study to assess the impact of three essential components in FedNed: the identification of extremely noisy clients (Id.), negative distillation (ND), and local pseudo-labeling (LPL).
The ablation study is carried out on CIFAR-10 with twenty clients including five EN clients (with noise ratio of 99\%).
The baseline without any component reverts to FedAvg which simply aggregates client models on the server.
With the inclusion of the identification of extremely noisy clients, we aggregate solely the identified MN clients for the global model.
When negative distillation is employed, the selected EN clients are utilized by optimizing the global model via Equation (\ref{negative_distillation}).
Regarding client-side training, if local pseudo-labeling is not adopted, we directly update the local model on its local dataset, regardless of the identification result on the server.
Conversely, in cases where local pseudo-labeling is employed, we update and upload two local models for each identified EN client, as described in Section \ref{step_local}.

Table \ref{ablation} shows the result of the ablation study. The most significant improvement is observed when identifying extremely noisy clients, supporting our conclusion that excluding them produces better outcomes than aggregating. Negative distillation results in an additional performance boost of approximately $2\%$, while local pseudo-labeling contributes around $1.5\%$. This validates the effectiveness of FedNed in handling extremely noisy clients. The highest performance is achieved when all components are used together.

\setlength{\tabcolsep}{12pt}
\begin{table}[!t]
	\centering                                    
	\begin{tabular}{cccc}                        
		\toprule                                  
		\textbf{Id.}&\textbf{ND}&\textbf{LPL}&\textbf{Acc.}\\          
		\midrule       
        \xmark &\xmark&\xmark&74.26\\
        \cmark &\xmark&\xmark&79.95\\
        \cmark &\cmark&\xmark&81.91\\
        \cmark &\xmark&\cmark&81.30\\
        \cmark &\cmark&\cmark&82.07\\

		\bottomrule                               
	\end{tabular}
 \caption{Ablation study of major components in FedNed.}
    \label{ablation}
    % \vspace{5px}
\end{table}

\subsubsection{Effectiveness of EN Client Identification via Uncertainty}\label{section_mc_dropout}
\begin{figure}[!t]
    \hspace{2px}
    \includegraphics[width=.9\linewidth]{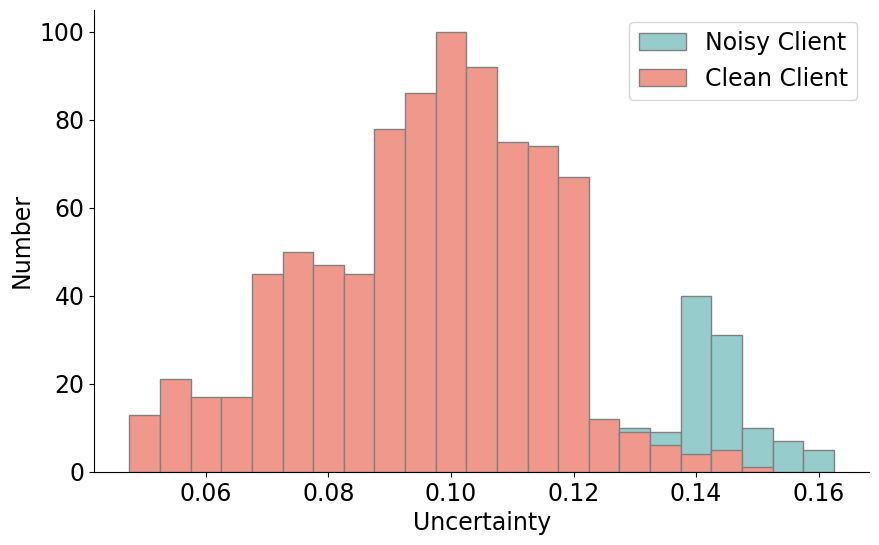}
    \caption{Histogram of model prediction uncertainty for both MN and EN clients, where the uncertainty is accumulated over all training rounds.}
    \label{mc_dropout_histogram}
\end{figure}

One key factor that leads to the success of FedNed is the accuracy of EN client identification.
In this regard, we evaluate the effectiveness of employing model prediction uncertainty with MC dropout as a distinguishing measure in FedNed to differentiate between MN and EN clients.
Figure \ref{mc_dropout_histogram} shows the histogram of model prediction uncertainty calculated by Equation (\ref{class_uncertainty}) for both MN and EN clients.
It can be observed that the uncertainty values significantly vary between MN and EN clients, allowing for the establishment of the threshold $\lambda$ within the range of $(0.12,0.14)$ for easy segregation.

\subsubsection{Effectiveness of Negative Distillation}
\begin{figure}[!t]
    \centering
    \begin{subfigure}[b]{0.48\linewidth}
        \includegraphics[width=1\linewidth]{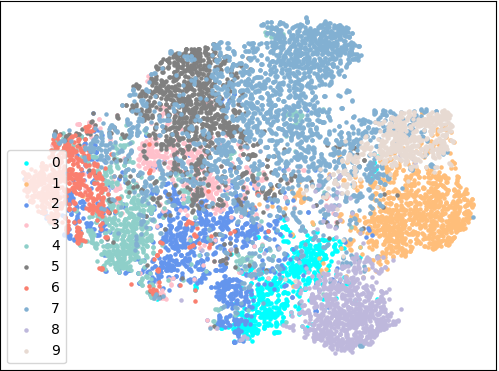}
        \caption{}
    \end{subfigure}
    \begin{subfigure}[b]{0.48\linewidth}
        \includegraphics[width=1\linewidth]{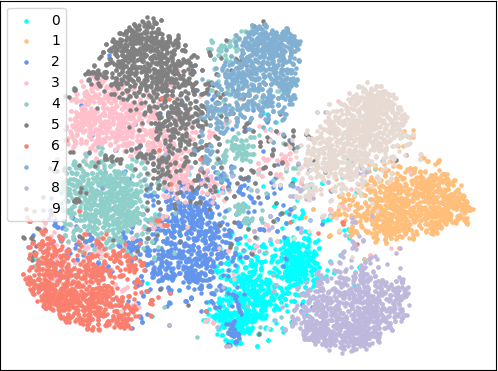}
        \caption{}
    \end{subfigure}
    \caption{Comparison in the feature spaces plotted with t-SNE. (a) FedNed without negative distillation on the server, (b) FedNed with negative distillation on the server.}
    \label{nd_tsne}
\end{figure}

In Section \ref{section_ablation}, we have shown the accuracy improvement of utilizing negative distillation to further improve the global model by incorporating the knowledge of EN client models.
In this context, we delve deeper into analyzing the nature of this improvement.
Figure \ref{nd_tsne} shows the t-SNE comparison between features of a global model before and after the employment of negative distillation trained on CIFAR-10.
% The visual representation clearly indicates that negative distillation aids in grouping features from the same class more coherently, as opposed to the one without negative distillation.
This affirms that the enhancement induced by negative distillation stems from an improved feature representation of the global model, aiding in more coherent grouping of similar types of features.

\subsubsection{Influence of the Number of Extremely Noisy Clients}
\begin{figure}[!t]
    \hspace{3px}
    \includegraphics[width=.9\linewidth]{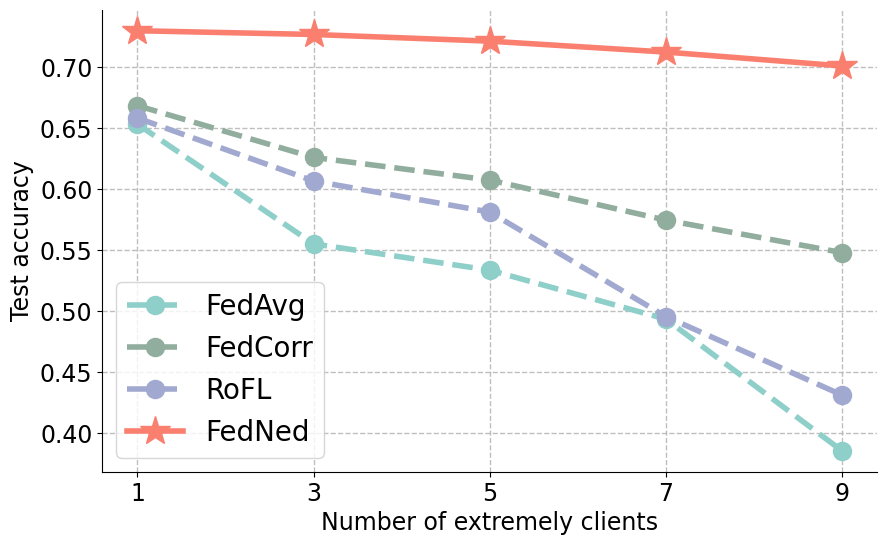}
    \caption{Comparison of performance among methods as the number of extreme noise clients increases.}
    \label{noise_client_lineplot}
\end{figure}

In this study, we have made an implicit assumption that the number of EN clients should be kept limited.
The rationale behind this assumption is that excessive EN clients could lead to a notable increase in the overall noise ratio across all clients, which could render the attainment of a robust global model unfeasible.
Therefore, in all previous experiments, we only set a few EN clients.
Nonetheless, an intriguing curiosity led us to investigate the performance of FedNed when the number of EN clients increases substantially, possibly even constituting up to half of the client population.
In this exploration, we set the total number of clients to twenty and vary the number of EN clients within the range of $[1, 3, 5, 7, 9]$.
Figure \ref{noise_client_lineplot} shows the performance degradation in accuracy as the number of EN clients increases, compared with all the other FL methods for label noise.
An unexpected observation emerges: despite the continuous increase in the number of extremely noisy (EN) clients, the accuracy of FedNed remains stable, while the accuracy of other methods experiences a significant decline.
It notably outperforms the baselines under these challenging circumstances.
This characteristic significantly positions FedNed as a potent solution to handling extreme scenarios where many noisy clients exist.

\subsubsection{Public Dataset Selection}
Considerations for further exploration include investigating the role of public datasets in FedNed. We believe that the effectiveness of the method is not dependent on the choice of public datasets. 
We've tried i) MNIST and ii) synthetic data generated by BigGAN adopted in FedDF. The results are slightly lower than when CIFAR-100 was used as the public dataset but still higher than the second-best method, FedNoRo. Table \ref{public_data} shows the results obtained using different public datasets, showcasing the adaptability of our proposed method.

\setlength{\tabcolsep}{12pt}
\begin{table}[!t]
	\centering                                    
	\begin{tabular}{cc}                        
		\toprule   
        \textbf{Public dataset}&\textbf{CIFAR-10}\\ 
		\midrule       
        CIFAR-100 & 84.97 \textbf{(2.56 $\uparrow$)} \\
        MNIST& 84.89 \textbf{(2.48 $\uparrow$)} \\
        Synthetic data & 83.60\textbf{ (1.19 $\uparrow$)}\\
		\bottomrule                               
	\end{tabular}
 \caption{The summary of the results with various public datasets shows the degree to which FedNed outperforms FedNoRo, as indicated in parentheses.}
    \label{public_data}
    % \vspace{5px}
\end{table}

\section{Concluding Remarks}
\paragraph{Conclusion} 
This paper addresses the critical challenge of handling noisy labels in federated learning (FL), especially in scenarios with highly contaminated clients experiencing extreme label noise. The proposed solution, FedNed, distinguishes extremely noisy clients and incorporates them into a knowledge distillation framework, optimizing their contributions. The negative distillation process, coupled with identification by MC dropout and local pseudo-labeling, enhances the trustworthiness of the global model from noisy clients while engaging all clients for aggregation. FedNed not only outperforms existing baselines but also establishes a new state-of-the-art in FL across diverse settings.

\paragraph{Limitations} Although the proposed FedNed mitigates performance degradation induced by extremely noisy clients in FL, a potential limitation lies in the degeneration of FedNed to FedAvg when no extremely noisy clients exist. One possible strategy involves treating FedNed as a plug-and-play module to identify extremely noisy clients, integrating with methods designed for handling mild label noise. 

\section{Acknowledgments}
This study was supported in part by the National Natural Science Foundation of China under Grants 62376233, U21A20514, 62006202 and 62376235; in part by the FuXiaQuan National Independent Innovation Demonstration Zone Collaborative Innovation Platform under Grant 3502ZCQXT2022008; in part by NSFC / Research Grants Council (RGC) Joint Research Scheme under Grant N\_HKBU214/21; and in part by the General Research Fund of RGC under Grants 12201321 and 12202622; in part by Guangdong Basic and Applied Basic Research Foundation No. 2022A1515011652.

\appendix
\bigskip
\bibliography{aaai24}
\clearpage
\section{Beta Distribution}
\begin{figure*}[ht]
    \centering
    \begin{subfigure}[b]{0.33\linewidth}
        \includegraphics[width=1\linewidth]{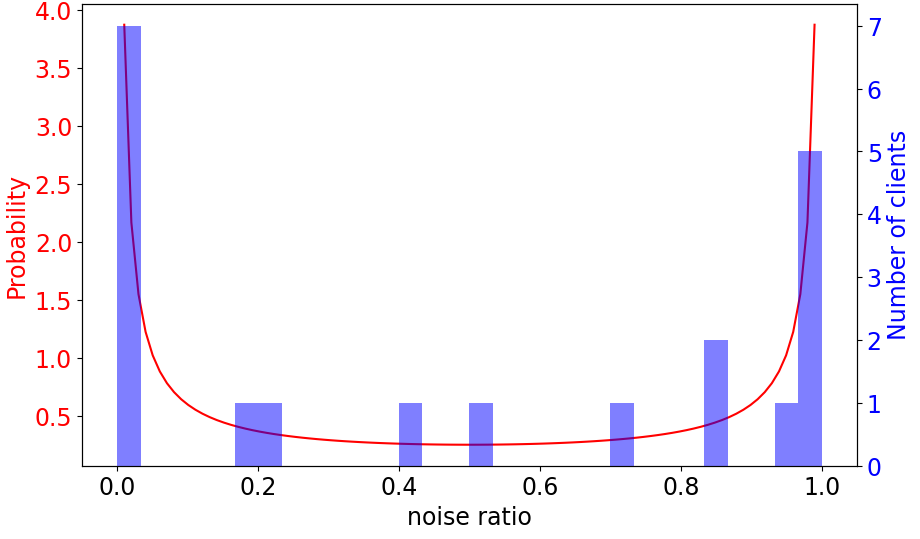}
        \caption{(0.1,0.1)}
    \end{subfigure}
    \begin{subfigure}[b]{0.33\linewidth}
        \includegraphics[width=1\linewidth]{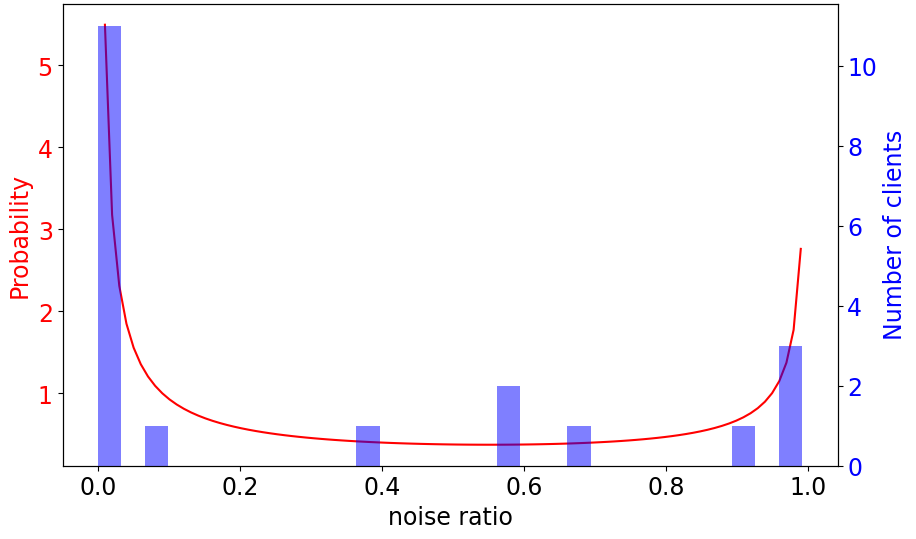}
        \caption{(0.1,0.3)}
    \end{subfigure}
    \begin{subfigure}[b]{0.33\linewidth}
        \includegraphics[width=1\linewidth]{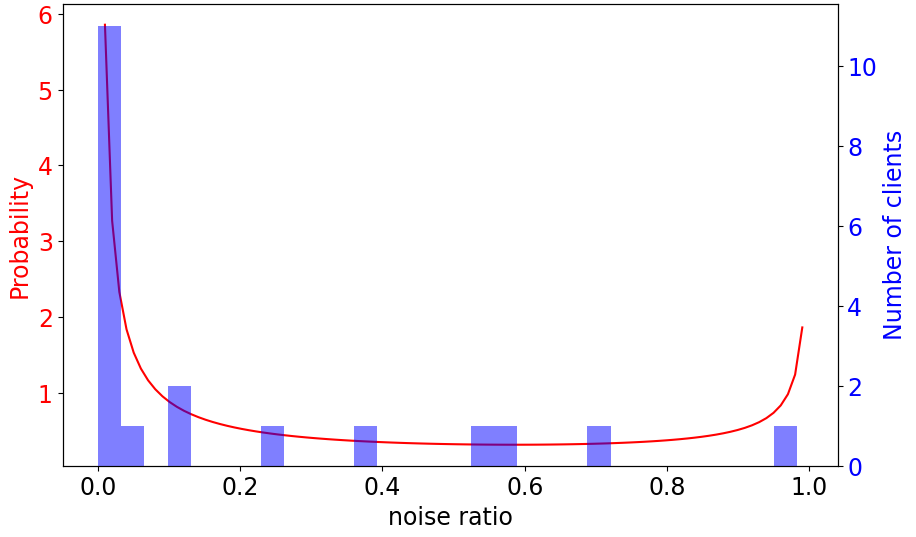}
        \caption{(0.3,0.5)}
    \end{subfigure}
    \caption{Noise ratio distribution drawn from Beta distribution from different parameters.}
    \label{beta}
\end{figure*}

In previous works, the configuration of noise ratios for noisy clients often followed a uniform distribution or a truncated Gaussian distribution. However, these samplings tended to concentrate the generated noise ratios around lower levels of noise, making it difficult to simulate scenarios with extreme noise. To better simulate extreme noisy client scenarios for comparison with state-of-the-art (SOTA) methods, we employed a setting based on sampling client noise ratios from a beta distribution. In our experiments involving 20 clients, we utilized three distinct beta distributions with parameters (0.1, 0.1), (0.1, 0.3), and (0.3, 0.5) to generate noise ratios. The resulting proportions and quantities of noisy clients are illustrated in Figure \ref{beta}.
The number of EN clients drawn from three Beta distributions (0.1, 0.1), (0.1, 0.3), and (0.3, 0.5) are about 5-6, 2-4, and 1-2, respectively, among a total number of 20 clients.
From these figures, we can observe that the sampling results of client noise ratios vary across different distributions. The overall noise ratio is approximately 20\%-30\%. 

\section{Addition Experimental Results}
\subsubsection{Learning with More Decentralized Clients and Less Participation Rate}
In our study, we delved into extremely noisy federated learning, utilizing a dataset with 20 clients and maintaining an active client ratio of 50\% in each round. However, federated learning systems often involve hundreds or even thousands of clients in total. To conduct a more comprehensive analysis, we conducted experiments on CIFAR10 with 100 clients, where the join ratio was set to 0.2, comparing our method with FedNoRo. The resulting accuracies were 66.81\% and 64.93\%, respectively, affirming the scalability of our approach. For this investigation, we chose a beta distribution of (0.1, 0.3), and the Dirichlet parameter was set at 0.7. Table \ref{B_a} shows comparative results with a higher number of decentralized clients and a lower participation rate.

\setlength{\tabcolsep}{12pt}
\begin{table}[ht]
	\centering                                    
	\begin{tabular}{ccc}                        
		\toprule   
        \textbf{Method}&\multicolumn{2}{c}{\textbf{CIFAR-10}}\\   
        \cmidrule{2-3} 
        Client number &  20 & 100\\
        \cmidrule{2-3} 
        Join ratio & 0.5 & 0.2\\
		\midrule       
        FedNoRo & 82.41 & 64.93\\
        FedNed & \textbf{84.97} & \textbf{66.81}\\

		\bottomrule                               
	\end{tabular}
 \caption{The results of learning with more decentralized clients and less participation rate.}
    \label{B_a}
    % \vspace{5px}
\end{table}

\subsubsection{Public Dataset Selection}
To delve deeper into the role of public datasets, in our experiments, we tested the method using i) MNIST and ii) synthetic data generated by BigGAN (as adopted in FedDF). The resulting accuracies on CIFAR-10 were $84.89\%$ and $83.60\%$, slightly lower than when CIFAR-100 was used as the public dataset but still surpassing the second-best method, FedNoRo. Table \ref{public_data} shows the results obtained using different public datasets, showcasing the adaptability of our proposed method.For this investigation, we chose a beta distribution of (0.1, 0.3), and the Dirichlet parameter was set at 0.7.

\subsubsection{Different Non-IID Levels}
We adhere to the methodology established in previous studies~\cite{xu2022fedcorr, yang2022robust} for setting benchmark datasets and conducting non-IID simulations. In our exploration, we further investigate our method with a non-IID coefficient $\alpha$ set to 0.1 on CIFAR-10. A larger $\alpha$ value results in less non-IID distributions for clients, meaning the distributions across different clients become more similar. In comparison with the second-best method, our approach consistently outperforms, with results on CIFAR-10 as follows: FedNoRo: $81.79\%$ and ours: $83.86\%$, showcasing the efficacy of our method. For this investigation, we chose a beta distribution of (0.1, 0.3). Table \ref{B_c} shows the results obtained at different non-IID levels.

\setlength{\tabcolsep}{12pt}
\begin{table}[ht]
	\centering                                    
	\begin{tabular}{cc}                        
		\toprule   
        \textbf{Method}&\textbf{CIFAR-10}\\  
        \midrule 
        Dirichlet & 0.1\\
		\midrule       
        FedNoRo & 81.67\\
        FedNed & \textbf{83.86}\\

		\bottomrule                               
	\end{tabular}
 \caption{The results of different non-IID levels.}
    \label{B_c}
    % \vspace{5px}
\end{table}

\section{Experiment Configeration Details}
\subsubsection{Comparison with SOTA Methods}
We use ResNet-18 for CIFAR-10, and ResNet-50 for CIFAR-100 as the base model. 
To ensure a fair comparison, all the compared FL methods are implemented with the same model architecture. 
All experiments are run by PyTorch on two NVIDIA GeForce RTX 3090 GPUs. 
By default, we run 100 communication rounds to present the experimental results.
The total number of clients is set at 20, and an active client ratio $50\%$ is maintained in each round. 
For local training, the batch size is set at 32. 
We use SGD with a learning rate $0.05$ as the optimizer.
For server negative distillation, we use Adam with a learning rate $0.05$ as the optimizer for negative distillation optimization processes.
The threshold $\lambda$ for the identification of EN client is set at $0.12$.
 The results are averaged from the top 10 global model accuracies achieved.

\subsubsection{Ablation Study}
We conduct an ablation study to assess the impact of three essential components in FedNed: the identification of extremely noisy clients (Id.), negative distillation (ND), and local pseudo-labeling (LPL).
The ablation study is carried out on CIFAR-10 with 20 clients. These comprised 15 clean clients (with a noise ratio of 0\%) and 5 EN clients (with a noise ratio of 99\%). The Dirichlet parameter was set at 0.7.

\subsubsection{Effectiveness of EN Client Identification via Uncertainty}
\begin{figure}[htb]
\centering
\includegraphics[width=.45\textwidth]{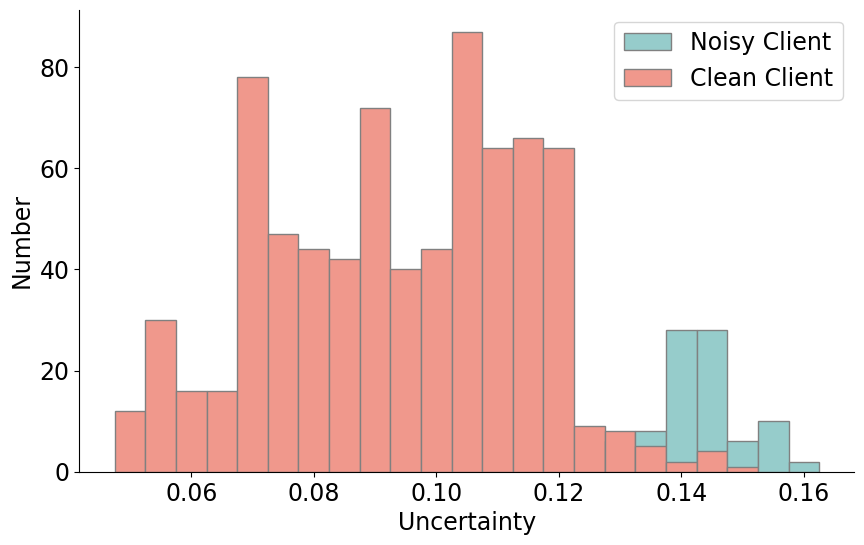} % Reduce the figure size so that it is slightly narrower than the column.
\caption{Histogram of model prediction uncertainty for both MN and EN clients, where the uncertainty is accumulated over all training rounds.In this figure,we employed the CIFAR-100 dataset and selected a beta distribution of (0.1, 0.3). The Dirichlet parameter was set at 0.7.}
\label{noise_client_lineplot_100}
\end{figure}

In this study, we employed the CIFAR-10/CIFAR-100 dataset and selected beta distribution of (0.1, 0.3). The Dirichlet parameter was set at 0.7. We computed the uncertainty of all participating clients over the course of 100 rounds. In this study, we made the assumption that the prior distribution of client noise ratios was known, enabling us to statistically determine the range of uncertainty values for both MN and EN.
It can be observed that the uncertainty values significantly vary between MN and EN clients, allowing for the establishment of the threshold $\lambda$ within the range of $(0.12,0.14)$ for easy segregation.
So we set $\lambda$ at 0.12 in all of other experiments. Figure \ref{noise_client_lineplot_100} illustrates the results obtained from the CIFAR-100 dataset.

\subsubsection{Effectiveness of Negative Distillation}
In this study, we delve deeper into analyzing the nature of this improvement.
we utilized the CIFAR-10 dataset with Dirichlet parameter set at 0.1. Due to the inherent randomness in beta distribution sampling, and in order to eliminate this uncertainty, we manually configured noise ratios for 20 clients. These comprised 15 clean clients (with a noise ratio of 0\%) and 5 EN clients (with a noise ratio of 99\%). Under the same configuration, we generated t-SNE plots for feature representations obtained through both negative distillation training for 80 rounds and standard aggregation for 80 rounds.

\subsubsection{Influence of the Number of Extremely Noisy Clients}
In this study, we have made an implicit assumption that the number of EN clients should be kept limited.
In all previous experiments, we only set a few EN clients, typically one or two.
Nonetheless, an intriguing curiosity led us to investigate the performance of FedNed when the number of EN clients increases substantially, possibly even dominating the client population.
In this exploration, we employed the CIFAR-10 dataset and set the Dirichlet parameter at 0.7. We manually controlled the number of EN clients to be $[1, 3, 5, 7, 9]$ with noise ratio of 99\% for EN clients and noise ratio of 0\% for the remaining clients.

\end{document}